\documentclass[letterpaper, 10 pt, journal, twoside]{IEEEtran}

\IEEEoverridecommandlockouts                              

\usepackage{bbm}
\usepackage{siunitx}
\usepackage{cite}
\usepackage{bm}
\usepackage{color}
\usepackage{graphbox,graphicx}
\usepackage{stfloats}
\usepackage{amsmath}
\usepackage{amssymb}
\usepackage{calligra}
\usepackage{bbm}
\usepackage{algorithmic}
\usepackage{array}
\usepackage{multirow}
\usepackage{tabularray}
\usepackage{makecell}
\usepackage{hhline}
\usepackage[utf8]{inputenc}
\usepackage{microtype}
\usepackage[caption=false,font=normalsize,labelfont=sf,textfont=sf]{subfig}
\usepackage{kotex}
\usepackage{float}
\usepackage{supertabular}
\usepackage[table]{xcolor} 
\usepackage{tabularx}
\usepackage{soul}
\usepackage{enumitem}
\usepackage{svg}
\usepackage{url}
\usepackage{booktabs}
\usepackage{threeparttable}
\usepackage[colorlinks=true, urlcolor=blue, linkcolor=red]{hyperref}
\usepackage{pifont}
\usepackage{dsfont}
\usepackage{amsmath, amssymb}

\usepackage{xcolor}

\title{HiPAN: Hierarchical Posture-Adaptive Navigation for Quadruped Robots in Unstructured 3D Environments}

\author{Jeil Jeong$^{1,*}$, Minsung Yoon$^{1,*}$, Seokryun Choi$^1$, Heechan Shin$^1$, Taegeun Yang$^1$, and Sung-eui Yoon$^{1,\dag}$%
\thanks{\phantom{Manuscript received: September 26, 2025; Revised December 29, 2025; Accepted April 13, 2026.                               
This paper was recommended for publication by Editor Abderrahmane Kheddar upon evaluation of the Associate Editor and Reviewers' comments.  
This work was supported by the Institute of Information \& Communications Technology Planning \& Evaluation (IITP) grant (No.               
RS-2025-25443318) and the National Research Foundation of Korea (NRF) grant (No. RS-2023-00208506) funded by the Korea government (MSIT).}} 
\thanks{$^*$These authors contributed equally to this work. $^1$All authors are with School of Computing, KAIST, South Korea. $^\dag$S. Yoon is a corresponding author; sungeui@kaist.edu.}%
}

\begin{document}

\maketitle

\begin{abstract}
Navigating quadruped robots in unstructured 3D environments poses significant challenges, requiring goal-directed motion, effective exploration to escape from local minima, and posture adaptation to traverse narrow, height-constrained spaces.
Conventional approaches employ a sequential mapping–planning pipeline but suffer from accumulated perception errors and high computational overhead, restricting their applicability on resource-constrained platforms.
To address these challenges, we propose Hierarchical Posture-Adaptive Navigation (HiPAN), a framework that operates directly on onboard depth images at deployment.
HiPAN adopts a hierarchical design: a high-level policy generates strategic navigation commands---planar velocity and body posture---which are executed by a low-level, posture-adaptive locomotion controller.
To mitigate myopic behaviors and facilitate long-horizon navigation, we introduce Path-Guided Curriculum Learning, which progressively extends the navigation horizon from reactive obstacle avoidance to strategic navigation.
In simulation, HiPAN achieves higher navigation success rates and greater path efficiency than classical reactive planners and end-to-end baselines, while real-world experiments further validate its applicability across diverse, unstructured 3D environments.
Videos of the experiments are available at \href{https://sgvr.kaist.ac.kr/~Jeil/project_page_HiPAN/}{project page}.
\end{abstract}


\section{Introduction} \label{sec:1}
\IEEEPARstart{Q}{uadruped} robots are increasingly deployed in unstructured three-dimensional (3D) environments---such as disaster zones and inspection sites---where dead-ends, cluttered layouts, and limited horizontal or vertical clearance are common~\cite{fan2021step, agha2021nebula}.
Navigating these environments demands not only strategic, goal-directed movement but also adaptive body posture adjustment, such as crouching or tilting, to traverse confined spaces and reach goal locations. 
While recent advances have demonstrated posture-adaptive locomotion for handling spatial constraints, most methods remain focused on reactive obstacle avoidance~\cite{zhuang2023robot, miki2024learning, han2024lifelike} or rely on manual commands~\cite{margolis2023walk, miao2025palo}, limiting their autonomy in unseen environments. 

For autonomous navigation, traditional approaches integrate mapping and path planning to generate locomotion commands for low-level controllers~\cite{fan2021step, agha2021nebula}.
However, explicit 3D mapping imposes computational and memory overhead on resource-constrained robotic platforms, requiring trade-offs between accuracy and real-time performance~\cite{buchanan2021perceptive, li2023autonomous, xu2024dexterous}. 
Alternatively, navigation strategies based on immediate local perception and relative goal position, such as Wall-Following~\cite{alamri2023autonomous} and BUG algorithms~\cite{zohaib2014improved, abafogi2018new, mcguire2019comparative}, offer computational and memory efficiency. 
However, their heuristic nature often produces inefficient paths and fails in cluttered, unstructured environments.

\begin{figure}[t!]
    \centering
    \includegraphics[width=\linewidth]{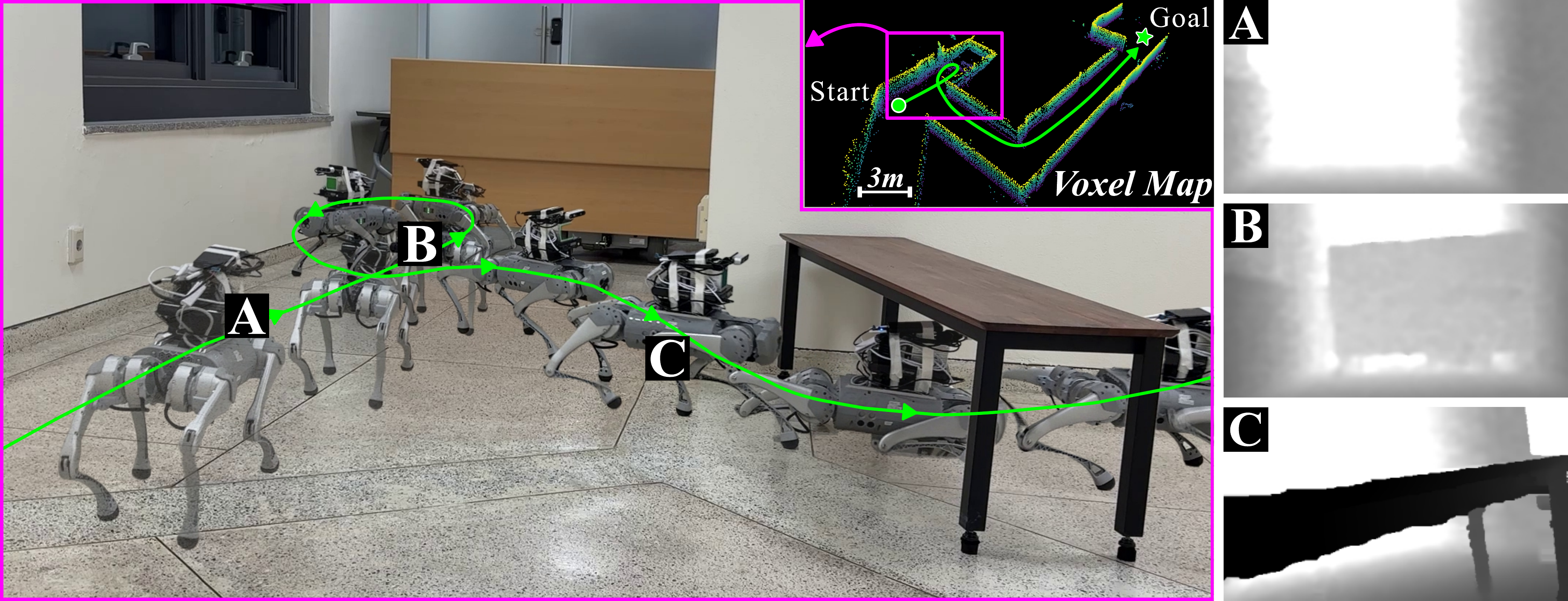}
    \vspace{-6.5mm}
    \caption{
    Our HiPAN framework enables long-horizon navigation in unstructured 3D environments, allowing a quadruped robot to backtrack after entering a dead-end corridor (A, B) and adaptively adjust its posture to pass under a height-constrained passage (C), relying solely on onboard depth perception during deployment. 
    The green curve indicates the robot's trajectory, with corresponding depth images from key moments (A,B,C) on the right.
    A voxel map is shown to illustrate the navigation scenario.
    }
    \label{fig:1}
    \vspace{-5.5mm}
\end{figure}

Reinforcement Learning (RL) has emerged as a promising approach for goal-directed navigation based on local perception, reducing the reliance on hand-crafted rules~\cite{hoeller2021learning, kareer2023vinl, he2024agile, rudin2022advanced, lee2024learning}.
However, in the absence of structured guidance during training, agents often exhibit myopic behavior, becoming trapped in local minima such as dead-end corridors or enclosed rooms~\cite{ao2021co, jang2021hindsight, gao2023efficient, gao2025hierarchical}.
Meanwhile, existing methods with structured learning schemes mainly focus on planar navigation, limiting their applicability in unstructured 3D environments where body posture adaptation is required to traverse confined spaces.
To address these limitations, we propose the Hierarchical Posture-Adaptive Navigation (HiPAN) framework that enables long-horizon navigation in unstructured 3D environments, as illustrated in Fig.~\ref{fig:1}.
Our main contributions are as follows:

\begin{itemize}[leftmargin=*]
\item We introduce the HiPAN framework, in which a high-level policy processes depth images to generate navigation commands---planar velocity and body posture---executed by a low-level locomotion policy.
\item We present Path-Guided Curriculum Learning (PGCL), which progressively extends the horizon during high-level policy training to mitigate myopic behavior and induce effective exploration strategies for long-horizon navigation.
\item We validate the HiPAN framework through extensive simulation experiments, achieving consistently high average success rates and path efficiency $(96.5\%,\ 88.7)$ compared to classical $(56.1\%,\ 41.7)$ and end-to-end baselines $(53.7\%,\ 48.4)$.
\item We demonstrate the robustness and practical applicability of our approach through real-world deployments across diverse obstacle configurations and lighting conditions.
\end{itemize}

\section{Related Works} \label{sec:2}
\subsection{Posture-Adaptive Locomotion of Quadruped Robots}
\label{sec:2-A}
Reinforcement learning (RL) has advanced quadrupedal locomotion over complex terrains~\cite{lee2020learning, miki2022learning, choi2023learning, kumar2021rma}. 
To enhance control flexibility, recent studies have expanded the conventional command space beyond the forward, lateral, and yaw velocities to include postural parameters like body height, roll, and pitch~\cite{margolis2023walk, miao2025palo}.
Exteroceptive inputs like depth images have been incorporated to adapt these parameters autonomously, enabling robots to traverse narrow passages and pass under overhanging obstacles~\cite{zhuang2023robot, miki2024learning, han2024lifelike}. 
However, these methods remain focused on immediate obstacle avoidance or traversal, while navigation-level velocity commands still depend on manual input or pre-defined paths, thereby limiting their autonomy in dynamic or previously unseen environments.
To address this, our HiPAN framework employs a high-level policy that autonomously generates both velocity and postural commands directly from local perception and goal information.
These commands are then executed by a posture-adaptive low-level locomotion policy, enabling robust, goal-directed traversal through confined spaces in unstructured 3D environments.

\vspace{-1mm}
\subsection{Policy Learning for Long-Horizon Navigation}
\label{sec:2-B}
Conventional navigation systems rely on environment mapping, path planning, and trajectory tracking via a locomotion controller~\cite{buchanan2021perceptive, li2023autonomous, xu2024dexterous}.
To simplify this pipeline, recent RL-based approaches employ policies that directly infer navigation commands from goal locations and local sensory inputs~\cite{hoeller2021learning, kareer2023vinl, he2024agile}.
However, relying solely on local observations renders these policies short-sighted.
Consequently, when optimized solely with distance-to-goal rewards, they often exhibit myopic behavior that prioritizes immediate progress, thereby becoming trapped in local minima---particularly in dead-end corridors or enclosed rooms common in unstructured environments.

State-based intrinsic rewards~\cite{tang2017exploration, kayal2025impact}, such as novelty, encourage broad exploration and help escape local minima, but lack the task-specific guidance required for long-horizon navigation in unstructured environments.
To address this, recent studies decompose the navigation problem into a sequence of short-horizon subproblems defined by intermediate subgoals derived from graph search or expert policies, directing exploration toward goals~\cite{ao2021co, jang2021hindsight, gao2023efficient, gao2025hierarchical}. 
Building on this, we combine intrinsic rewards and subgoal-based exploration as learning strategies to develop an effective high-level policy for long-horizon navigation. 
Combined with posture adaptability, we extend prior studies limited to 2D floor plans into unstructured 3D environments, thereby enabling robots to escape local minima and safely traverse confined spaces using local perception.

\vspace{-1mm}
\subsection{Hierarchical Frameworks}
\label{sec:2-C}
Hierarchical frameworks decompose complex robotic tasks into modular sub-tasks, allowing each layer to be optimized using domain-specific knowledge~\cite{feng2024learning, zhang2024resilient, seo2023learning}.
In navigation, the high-level policy processes sensory inputs to generate velocity commands that remain within the control capabilities of the low-level policy, which translates them into executable motor actions~\cite{hoeller2021learning, kareer2023vinl}.
This modular design offers key advantages: the low-level policy can be robustly pre-trained and reused across tasks, while the separation improves training efficiency and the interpretability of each component~\cite{seo2023learning, lee2024learning}.
Therefore, we adopt a hierarchical framework to address complex navigation challenges in unstructured 3D environments, utilizing the posture-adaptive capabilities of quadruped robots.

\begin{figure}[t!]
    \centering
    \includegraphics[width=\linewidth]{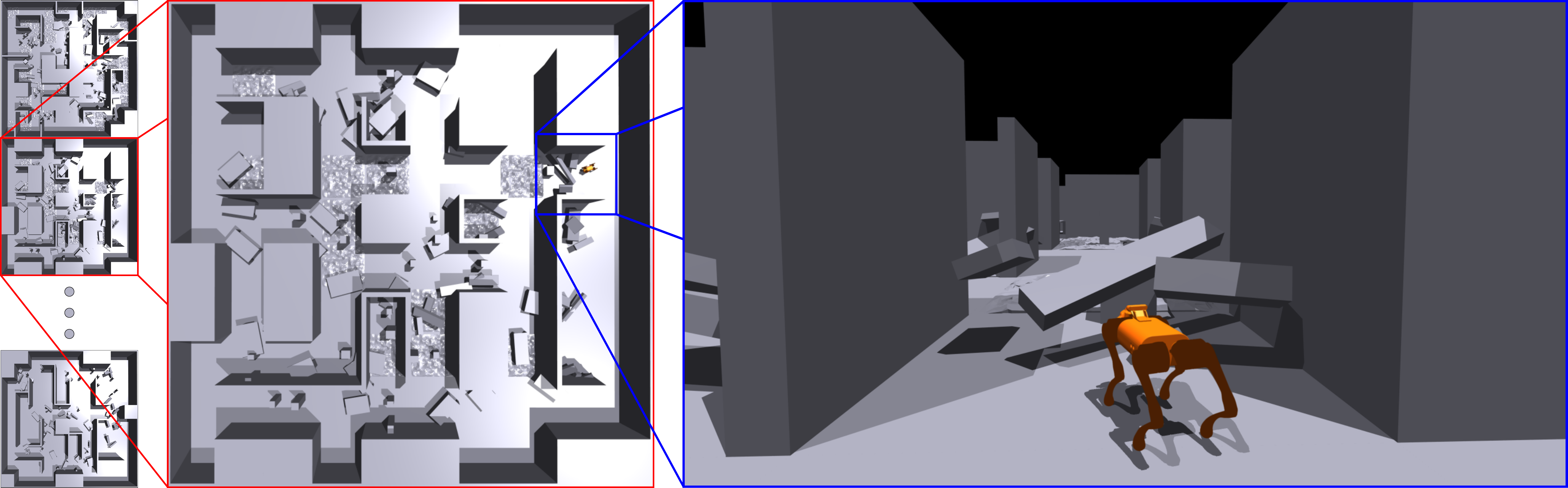}
    \vspace{-7mm}
    \caption{Representative examples of unstructured 3D environments generated via the Wave Function Collapse (WFC) algorithm. 
    See Sec.~\ref{sec:task-formulation} for details.}
    \label{fig:env_examples}
    \vspace{-6mm}
\end{figure}

\section{Variable Notation}
We represent Cartesian position, linear velocity, and linear acceleration as $\boldsymbol{p}$, $\boldsymbol{v}$, and $\dot{\boldsymbol{v}} \in \mathbb{R}^3$, respectively. 
Orientation is denoted by XYZ Euler angles $\boldsymbol{\theta}$, with angular velocity $\boldsymbol{\omega}$ in $\mathbb{R}^3$.
The quadruped robot has twelve joints, whose position, velocity, acceleration, and torque are denoted by $\boldsymbol{q}$, $\dot{\boldsymbol{q}}$, $\ddot{\boldsymbol{q}}$, and $\boldsymbol{\tau} \in \mathbb{R}^{12}$, respectively. 
Each foot $F_i$ ($i \in \{0,1,2,3\}$) has a contact state represented by a binary indicator $I_i \in \{0,1\}$.
For clarity, superscripts denote reference frames (world $\mathcal{W}$ or body $\mathcal{B}$), and subscripts indicate entities, components, or time; for example, $v^{\mathcal{W}}_{B, z, t}$ denotes the $z$-component of the robot's body ($B$) velocity in the world frame at time step $t$. 
The component and time index are omitted when unnecessary for brevity.

\section{Methodology}\label{sec:method}
We propose the Hierarchical Posture-Adaptive Navigation (HiPAN) framework, which employs a two-level hierarchy of reinforcement learning policies optimized in a stage-wise manner.
The high-level policy processes local perception to generate strategic navigation commands, enabling the robot to escape local minima (\textit{e.g.}, dead-end corridors) and traverse confined spaces.
The low-level policy executes these commands via coordinated joint motions, adaptively adjusting body posture.
The subsequent sections describe the task and environment formulation, policy architectures and training paradigms at each level, and implementation details.

\subsection{Task and Environment Formulation} \label{sec:task-formulation}
We formulate the task as navigating a quadruped robot to a goal position $\boldsymbol{g} \in \mathbb{R}^3$ within an unstructured 3D environment. 
The environment presents challenges such as uneven surfaces, overhanging structures requiring posture adaptation, and local trap regions (\textit{e.g.}, dead-end corridors, semi-enclosed rooms) that demand non-greedy navigation strategies.
To compose such complex environments effectively, we employ the Wave Function Collapse algorithm~\cite{Gumin_Wave_Function_Collapse_2016, miki2024learning}, a tile-based method that assembles terrain by enforcing local adjacency constraints. 
The tiles are comprised of flat, rough, and obstructed surfaces, stochastically arranged to create diverse layouts. 
In addition, we randomly place walls, tables, and floating boxes to create local trap regions and overhanging obstacles.
Fig.~\ref{fig:env_examples} shows procedurally generated task environments.

\begin{figure*}[ht]
\centering
\includegraphics[align=t, width=\textwidth]{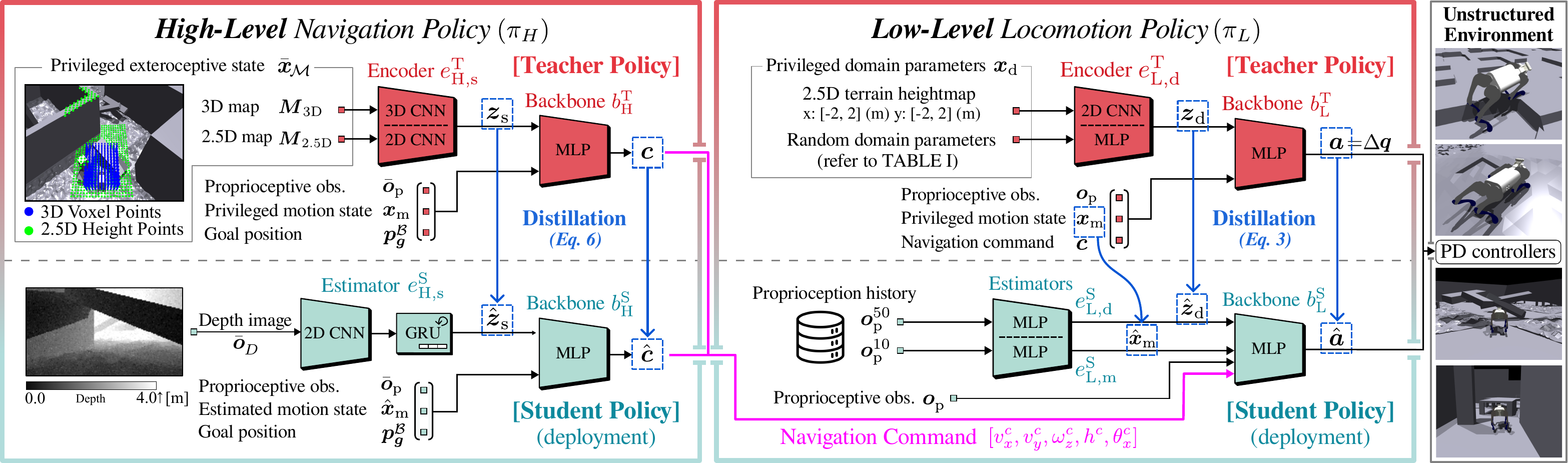}
\vspace{-2mm}
\caption{
\textbf{Overview of the proposed HiPAN framework.} Teacher policies (red) are trained with privileged inputs via Proximal Policy Optimization~\cite{schulman2017proximal}, and student policies (cyan) are distilled using Dataset Aggregation~\cite{ross2011reduction} with only onboard sensory data. 
During deployment, only student policies are used to navigate toward the goal in unstructured 3D environments.
Encoders and estimators bridge the observation gap between simulation and the real world.
}
\label{fig:framework}
\vspace{-6mm}
\end{figure*}

\subsection{Low-Level Locomotion Policy with Posture Adaptation} \label{sec:low-level-policy}
As illustrated in Fig.~\ref{fig:framework}, the low-level policy $\pi_{\text{L}}$ maps high-level navigation commands $\boldsymbol{c}$ into whole-body joint actions $\boldsymbol{a} \in \mathbb{R}^{12}$. 
It serves as the physical execution layer within the HiPAN framework, ensuring that high-level decisions $\boldsymbol{c}$ are realized with robustness to physical variations (\textit{e.g.}, uneven surfaces, sim-to-real domain gaps) and external perturbations.

\subsubsection{Navigation Command}
The navigation command $\boldsymbol{c} = [v_{x}^{c}, v_{y}^{c}, \omega_{z}^{c}, h^{c}, \theta_{x}^{c}] \in \mathcal{C} \subset \mathbb{R}^5$ is a five-dimensional vector, where the first three elements specify the desired forward, lateral, and yaw velocities of the body in the body frame, controlling planar motion. 
The last two define the target body height relative to the ground and roll angle in the world frame for postural adjustments to traverse confined spaces.
This compact yet expressive interface facilitates communication between the high-level and low-level policies for long-horizon navigation in unstructured 3D environments.

\subsubsection{Policy Learning}
We develop the low-level locomotion policy using a two-stage teacher-student framework to enable robust sim-to-real transfer~\cite{lee2020learning, kumar2021rma, chen20learning}.

In the first stage, we train the teacher policy $\pi_{\text{L}}^{\text{T}} : \boldsymbol{s} \rightarrow \boldsymbol{a}$ via Reinforcement Learning (RL) in a fully observable Markov Decision Process (MDP) defined by the tuple $(\mathcal{S}, \mathcal{A}, \mathcal{P}, \mathcal{R}, \rho_0, \gamma)$, which specifies state and action spaces, transition dynamics, reward function, initial state distribution, and discount factor.
At the start of each episode, we sample $\boldsymbol{s}_0 \sim \rho_0$ by placing the robot on terrain in the nominal standing posture $\boldsymbol{q}_0$.
We then optimize the teacher-policy parameters $\phi_{\text{L}}^{\text{T}}$ to maximize the expected return over the command distribution $P(\boldsymbol{c})$:
\begin{equation}
\pi_{\text{L}}^{\text{T}} := \pi_{\text{L}}^{\text{T}}(\cdot;\phi_{\text{L}}^{\text{T}*}), \quad \phi_{\text{L}}^{\text{T}*} = \arg\max_{\phi_{\text{L}}^{\text{T}}} J(\phi_{\text{L}}^{\text{T}}),
\end{equation}
\vspace{-2mm}
\begin{equation}
J(\phi_{\text{L}}^{\text{T}}) = 
\mathbb{E}_{\boldsymbol{c} \sim P(\boldsymbol{c})} \!
\left[
\mathbb{E}_{\substack{(\boldsymbol{s}, \boldsymbol{a}) \sim \rho_{\phi_{\text{L}}^{\text{T}}} \\ \boldsymbol{s}_0 \sim \rho_0}} \!
\left[ 
\sum_{t=0}^{\infty} \gamma^{t} \mathcal{R}(\boldsymbol{s}_t, \boldsymbol{a}_t \! \mid \boldsymbol{c})
\right]
\right],
\label{eq:objective_function}
\end{equation}
where $\rho_{\phi_{\text{L}}^{\text{T}}}$ is the state-action visitation distribution under the parameter $\phi_{\text{L}}^{\text{T}}$.
To cover a broad range of the command space $\mathcal{C}$, we employ the grid-adaptive curriculum~\cite{margolis2024rapid}, which gradually expands $P(\boldsymbol{c})$ by introducing higher planar velocity commands once postural commands$(h^{c},\theta_{x}^{c})$ are reliably tracked.

In the second stage, we train the student policy $\pi_{\text{L}}^{\text{S}}: \boldsymbol{o} \rightarrow \hat{\boldsymbol{a}}$ under partial observability, using only on-board sensory inputs as observations $\boldsymbol{o} \in \mathcal{O} \subset \mathcal{S}$. 
We distill the student $\pi_{\text{L}}^{\text{S}}$ from the optimized teacher $\pi_{\text{L}}^{T}$ via the Dataset Aggregation (DAgger) algorithm~\cite{ross2011reduction}.
The goal is to imitate the teacher's action $\boldsymbol{a}$ and estimate the privileged information provided by the simulator, such as domain latent vector $\boldsymbol{z}_{\text{d}}$ and motion states $\boldsymbol{x}_{\text{m}}$ as shown in Fig.~\ref{fig:framework}.
To this end, we optimize the student-policy parameters $\phi_{\text{L}}^{\text{S}}$ by minimizing the following loss: 
\begin{align}
\mathcal{L}_{\text{DAgger}}(\phi_{\text{L}}^{\text{S}}) = 
& \mathbb{E}_{\mathcal{D}} \big[ 
\left\| \boldsymbol{a} - \hat{\boldsymbol{a}}(\phi_{\text{L}}^{\text{S}}) \right\|_2^2 + \notag \\
& \left\| \boldsymbol{z}_{\text{d}} - \hat{\boldsymbol{z}}_{\text{d}}(\phi_{\text{L}}^{\text{S}}) \right\|_2^2
 + \left\| \boldsymbol{x}_{\text{m}} - \hat{\boldsymbol{x}}_{\text{m}}(\phi_{\text{L}}^{\text{S}}) \right\|_2^2 \big],
\end{align}
where $\hat{\boldsymbol{a}}$, $\hat{\boldsymbol{z}}_{\text{d}}$, and $\hat{\boldsymbol{x}}_{\text{m}}$ are the student’s predictions, parameterized by $\phi_{\text{L}}^{\text{S}}$.
For the dataset $\mathcal{D}$, we iteratively roll out the student and annotate its observations with teacher supervision.

\subsubsection{Policy Architecture}
Both teacher and student policies use identical backbone architectures $b_{\text{L}}^{\{\text{T}, \text{S}\}}\!$, which output joint displacements as actions $\boldsymbol{a} = \Delta \boldsymbol{q}$. 
Proportional-derivative controllers then compute torques $\boldsymbol{\tau}$ to track the targets $\boldsymbol{q}_0 + \Delta\boldsymbol{q}$.

The teacher $\pi_{\text{L}}^{\text{T}}$ receives a 4-tuple state $\boldsymbol{s} = (\boldsymbol{o}_{\text{p}}, \boldsymbol{c}, \boldsymbol{x}_{\text{m}}, \boldsymbol{x}_{\text{d}})$ as policy inputs, consisting of:
(1) a proprioceptive observation $\boldsymbol{o}_{\text{p}} = [\boldsymbol{q}, \dot{\boldsymbol{q}}, (\boldsymbol{p}_{F_{0}}^{\mathcal{B}}, \dots, \boldsymbol{p}_{F_{3}}^{\mathcal{B}}), (I_{0}, \dots, I_{3}), \boldsymbol{\theta}_{B, xy}^{\mathcal{W}}, \boldsymbol{\omega}_B^{\mathcal{B}}, \boldsymbol{a}_{t-1}]$, including joint positions and velocities, foot positions and contact indicators, body orientation in roll and pitch, angular velocity, and the previous action;
(2) the navigation command $\boldsymbol{c}$;
(3) privileged motion states $\boldsymbol{x}_{\text{m}} = [\boldsymbol{v}_B^{\mathcal{B}}, h_{B}, \theta_{B,x}^{\mathcal{W}}]$, comprising the body linear velocity in the body frame, and the body height relative to the ground and roll angle in the world frame; and
(4) privileged domain parameters $\boldsymbol{x}_{\text{d}}$, including local terrain heightmaps and domain-randomized variables such as friction (refer to TABLE~\ref{table:dom-rnd} for details).
To enhance adaptability for physical variations, a domain-parameter encoder $e_{\text{L}, \text{d}}^{\text{T}}$ embeds $\boldsymbol{x}_{\text{d}}$ into the latent vector $\boldsymbol{z}_{\text{d}} \in \mathbb{R}^{32}$, which is concatenated with the other inputs and passed to $b_{\text{L}}^{\text{T}}: \boldsymbol{o}_{\text{p}} \times \boldsymbol{c} \times \boldsymbol{x}_{\text{m}} \times \boldsymbol{z}_{\text{d}} \rightarrow \boldsymbol{a}$.

On the other hand, the student policy $\pi_{\text{L}}^{\text{S}}$ takes as inputs the observation $\boldsymbol{o} = [\boldsymbol{o}_{\text{p}}^{50}, \boldsymbol{c}]$, where $\boldsymbol{o}_{\text{p}}^{50} = \{\boldsymbol{o}_{\text{p}, t}, \dots, \boldsymbol{o}_{\text{p}, t - 49}\}$ denotes a history of proprioceptive observations and $\boldsymbol{c}$ is the command.
To address the absence of privileged information during deployment, we introduce two estimators: a domain estimator $e_{\text{L}, \text{d}}^{\text{S}}: \boldsymbol{o}_{\text{p}}^{50} \rightarrow \hat{\boldsymbol{z}}_{\text{d}}$ for latent domain features $\boldsymbol{z}_{\text{d}}$ and a motion estimator $e_{\text{L}, \text{m}}^{\text{S}}: \boldsymbol{o}_{\text{p}}^{10} \rightarrow \hat{\boldsymbol{x}}_{\text{m}}$ for motion states $\boldsymbol{x}_{\text{m}}$.
The predicted and observed features are concatenated and passed to the student backbone network $b_{\text{L}}^{\text{S}}: \boldsymbol{o}_{\text{p}} \times \boldsymbol{c} \times \hat{\boldsymbol{x}}_{\text{m}} \times \hat{\boldsymbol{z}}_{\text{d}} \rightarrow \hat{\boldsymbol{a}}$.

\subsection{High-Level Navigation Policy} \label{sec:high-level-policy}
The high-level policy $\pi_{\text{H}}$ serves as the strategic navigation layer within our HiPAN framework. 
It generates the navigation commands $\boldsymbol{c}$ to guide the robot toward goals $\boldsymbol{g}$ in unstructured 3D environments by making decisions on posture adaptation for confined-space traversal, obstacle-avoidance maneuvers, and non-greedy exploration to escape local minima.

\begin{figure}[t!]
    \centering
    \includegraphics[width=\linewidth]{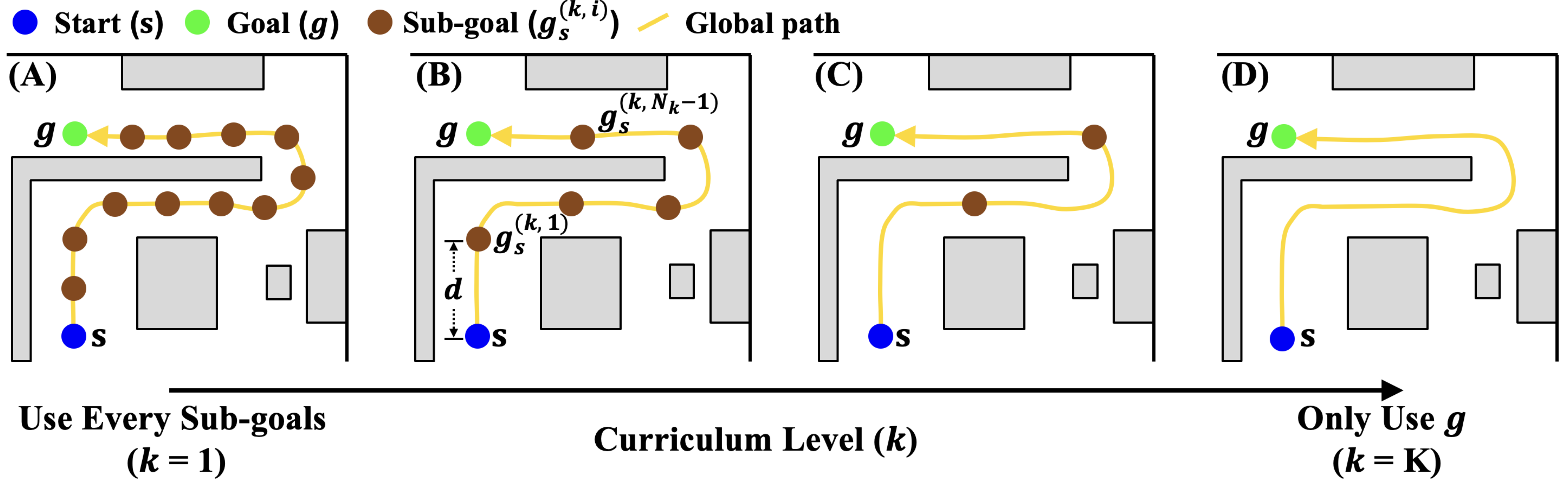}
    \vspace{-8mm}
    \caption{Path-guided curriculum learning progressively removes intermediate sub-goals along a privileged global path, gradually increasing navigation horizon and task difficulty. 
    In the early stages, the agent sequentially reaches every sub-goal (A), enabling it to first acquire reactive obstacle avoidance skills. 
    As the curriculum level increases, only a subset of sub-goals are provided (B, C), transitioning to direct navigation from start to goal (D).
    } 
    \label{fig:way}
    \vspace{-5mm}
\end{figure}

\subsubsection{Policy Learning}
The high-level policy $\pi_{\text{H}}$ is also trained via the teacher-student paradigm. 
The teacher policy $\pi_{\text{H}}^{\text{T}}$ is first optimized with privileged information using RL, and its behavior and spatial-perceptual latent representations are distilled into the student policy $\pi_{\text{H}}^{\text{S}}$ via the DAgger algorithm. 
This procedure enables the student to rely solely on onboard depth perception during deployment, eliminating the need for privileged map information, as shown in Fig.~\ref{fig:framework}.

We formulate the teacher's navigation problem as a fully observable MDP represented by the tuple $(\bar{\mathcal{S}}, \bar{\mathcal{A}}, \bar{\mathcal{P}}, \bar{\mathcal{R}}, \bar{\rho}_0, \bar{\gamma})$, consisting of high-level state and action spaces, transition dynamics, reward function, initial state distribution, and discount factor.
The notation $\bar{(\cdot)}$ is used to indicate arbitrary elements specific to the high-level MDP.
As illustrated by the magenta-dashed line in Fig.~\ref{fig:framework}, we embed the low-level student policy $\pi_{\text{L}}^{\text{S}}$ into the high-level transition dynamics $\bar{\mathcal{P}}$ to reflect the hierarchical control structure, capturing closed-loop effects such as command-tracking errors and execution delays. 
The high-level policy $\pi_{\text{H}}$ produces the navigation commands as high-level actions $\bar{\boldsymbol{a}} = \boldsymbol{c} \in \bar{\mathcal{A}}$. 
For navigation problems, we uniformly sample $\bar{\boldsymbol{s}}_0 \sim \bar{\rho}_0$ and $\boldsymbol{g} \sim P(\boldsymbol{g})$ by selecting collision-free start and goal positions, placing the robot in the standing posture $\boldsymbol{q}_0$.
We then optimize the teacher-policy parameters $\phi_{\text{H}}^{\text{T}}$ by maximizing the expected return under the start $\bar{\rho}_0$ and goal $P(\boldsymbol{g})$ distributions:
\begin{equation}
\pi_{\text{H}}^{\text{T}} := \pi_{\text{H}}^{\text{T}}(\cdot;\phi_{\text{H}}^{\text{T}*}), \quad \phi_{\text{H}}^{\text{T}*} = \arg\max_{\phi_{\text{H}}^{\text{T}}} J(\phi_{\text{H}}^{\text{T}}), 
\end{equation} 
\vspace{-2mm}
\begin{equation}
J(\phi_{\text{H}}^{\text{T}}) =
\mathbb{E}_{\substack{
    \boldsymbol{g} \sim P(\boldsymbol{g}) \\ \bar{\boldsymbol{s}}_0 \sim \bar{\rho}_0}} \!
\left[
\mathbb{E}_{(\bar{\boldsymbol{s}},\, \boldsymbol{c}) \sim \bar{\rho}_{\phi_{\text{H}}^{\text{T}}}} \!
\left[
\sum_{t=0}^{\infty} \bar{\gamma}^{t} \, \bar{\mathcal{R}}(\bar{\boldsymbol{s}}_t,\, \boldsymbol{c}_t \mid \boldsymbol{g})
\right]
\right].
\label{eq:high-level-objective}
\end{equation}

After training the teacher policy $\pi_{\text{H}}^{\text{T}}$, the student $\pi_{\text{H}}^{\text{S}}$ collects depth-image observations through interaction with the environment.
The teacher annotates these observations with the corresponding command $\boldsymbol{c}$ and spatial–perceptual latent feature $\boldsymbol{z}_{\text{s}}$.
The student is then trained on the aggregated dataset $\bar{\mathcal{D}}$ by minimizing the following regression loss, where $\hat{\boldsymbol{c}}$ and $\hat{\boldsymbol{z}}_{\text{s}}$ denote the student's predictions parameterized by $\phi_{\text{H}}^{\text{S}}$.
\begin{equation}
\mathcal{L}_{\text{DAgger}}(\phi_{\text{H}}^{\text{S}}) = 
\mathbb{E}_{\bar{\mathcal{D}}} 
\big[
\left\| \boldsymbol{c} - \hat{\boldsymbol{c}}(\phi_{\text{H}}^{\text{S}}) \right\|_2^2 +
\left\| \boldsymbol{z}_{\text{s}} - \hat{\boldsymbol{z}}_{\text{s}}(\phi_{\text{H}}^{\text{S}}) \right\|_2^2 
\big].
\end{equation}

\subsubsection{Path-Guided Curriculum Learning} \label{sec:way}
Learning effective navigation in complex, unstructured environments is challenging, as it requires balancing goal-directed movement with exploration to escape local minima using only local perception.
When policies are optimized solely on distance-to-goal rewards, they often learn a myopic behavior, greedily pursuing the goal and frequently getting trapped in local minima~\cite{ao2021co, jang2021hindsight, gao2023efficient, gao2025hierarchical}.

To mitigate this issue, we incorporate an Intrinsic Reward (IR) via a State Count mechanism~\cite{strehl2008analysis} to encourage visiting novel states (refer to $r^H_2$ in TABLE~\ref{table:high-level-rwd}). 
While this promotes state exploration~\cite{kayal2025impact}, we observe that this undirected exploration lacks the structured, task-specific guidance needed for efficient long-horizon navigation in unstructured environments.

To introduce this structured guidance, we propose a strategy, \textit{path-guided curriculum learning}, which decomposes a navigation task into progressively harder sub-goal-reaching tasks as illustrated in Fig.~\ref{fig:way}. 
At each level $k$, the agent receives a set of intermediate sub-goals between the start and goal locations:
\begin{equation}
    \mathcal{G}_k = \{\boldsymbol{g}_s^{(k, 1)}, \dots, \boldsymbol{g}_s^{(k, N_k-1)}, \boldsymbol{g}_s^{(k, N_k)} \! := \boldsymbol{g} \}, 
\end{equation}
where $N_k$ denotes the number of sub-goals, and the last sub-goal is the original goal $\boldsymbol{g}$.
As depicted in Fig.~\ref{fig:way}, sub-goals are sampled every $d$ meters along a global path computed with privileged map information.
We begin with $k = 1$ and $d = \SI{1}{m}$, and increase level $k$ by $\SI{1}{m}$ increments of $d$ each time the agent sequentially reaches all $N_k$ sub-goals. 
At any point during level $k$, the policy's goal observation becomes the next unreached sub-goal in the sequence $\mathcal{G}_k$.
This continues until $d$ exceeds the path length, which defines the final level~$K$ where only the original goal remains: $\mathcal{G}_K = \{\boldsymbol{g}_s^{(K, 1)} := \boldsymbol{g}\}$.
Once the agent consistently succeeds at level~$K$, we assign a new navigation problem with randomly sampled start and goal locations.
This curriculum progressively expands the agent's navigation horizon, enabling it to first acquire reactive skills—obstacle avoidance and confined-space traversal—and later develop well-balanced navigation strategies for exploration and goal-directed progress.
Please note that maps and sub-goals are used only during teacher training as privileged information, while the student operates under partial observability at deployment.

Formally, we extend the high-level objective function in Eq.~(\ref{eq:high-level-objective}) by optimizing the teacher policy to maximize the expected return over all curriculum levels $k$ and associated sub-goals $i$. 
The extended objective $J_{\text{PGCL}}(\phi_{\text{H}}^{\text{T}})$ is defined as:
\begin{equation}
    J_{\text{PGCL}}(\phi_{\text{H}}^{\text{T}}) = 
    \mathbb{E}_{\substack{\boldsymbol{g} \sim P(\boldsymbol{g}) \\ \bar{\boldsymbol{s}}_0 \sim \bar{\rho}_0}} \!
    \left[
    \sum_{k=1}^{K} \sum_{i=1}^{N_k} J_{k,i}(\phi_{\text{H}}^{\text{T}})
    \right],
\end{equation}
where $J_{k,i}$ denotes the expected return for sub-goal $i$ at level~$k$:
\begin{equation}
J_{k,i}(\phi_{\text{H}}^{\text{T}}) =
\mathbb{E}_{\substack{(\bar{\boldsymbol{s}}_t, \boldsymbol{c}_t) \sim \bar{\rho}_{\phi_{\text{H}}^{\text{T}}}^{(k,i)} \\
\bar{\boldsymbol{s}}_0^{(k,i)} \sim \bar{\rho}_0^{(k,i)}}
} \left[
\sum_{t=0}^{T_{k,i}} \bar{\gamma}^t \bar{\mathcal{R}}(\bar{\boldsymbol{s}}_t, \boldsymbol{c}_t \mid \boldsymbol{g}_s^{(k,i)})
\right],
\end{equation}
where $\bar{\rho}_{\phi_{\text{H}}^{\text{T}}}^{(k,i)}$ and $\bar{\rho}_0^{(k,i)}$ denote the state-action and initial state distributions of the MDP targeting the $i$-th sub-goal at level~$k$. 
Here, $\bar{\rho}_0^{(k,i)}$ is the state distribution upon reaching $\boldsymbol{g}_s^{(k,i-1)}$, and $T_{k,i}$ is the first timestep the agent reaches the sub-goal:
\begin{equation}
    T_{k,i} = \min\left\{t : \left\| \boldsymbol{p}_{B}^{\mathcal{W}} - \boldsymbol{p}_{\boldsymbol{g}_s^{(k,i)}}^{\mathcal{W}} \right\|_2 < \SI{0.1}{\meter} \right\}.
\end{equation}

\subsubsection{Policy Architecture}
The high-level teacher and student policies have the same backbone architecture $b_{\text{H}}^{\{\text{T}, \text{S}\}}$, issuing navigation commands $\bar{\boldsymbol{a}} = \boldsymbol{c}$ as actions, as shown in Fig.~\ref{fig:framework}.

The teacher $\pi_{\text{H}}^{\text{T}}$ takes a 4-tuple state $\bar{\boldsymbol{s}} = (\bar{\boldsymbol{x}}_{\!\mathcal{M}}, \bar{\boldsymbol{o}}_{\text{p}}, \boldsymbol{x}_{\text{m}}, \boldsymbol{p}_{\boldsymbol{g}_s}^{\mathcal{B}})$ as input, where
(1) $\bar{\boldsymbol{x}}_{\!\mathcal{M}} \!=\! (\boldsymbol{M}_{3\text{D}} \!\in \mathbb{R}^{14 \times 11 \times 11}, \boldsymbol{M}_{2.5\text{D}} \!\in \mathbb{R}^{31 \times 21})$ are the privileged 3D occupancy and 2.5D elevation local maps around the robot, sampled at \SI{0.1}{\meter} resolution; 
(2) $\bar{\boldsymbol{o}}_{\text{p}} = [\boldsymbol{o}_{\text{p}}, \boldsymbol{c}_{t-1}]$ is the high-level proprioceptive observation, consisting of the low-level's $\boldsymbol{o}_{\text{p}}$ and the previous command $\boldsymbol{c}_{t-1}$;
(3) $\boldsymbol{x}_{\text{m}}$ is the privileged motion state; and 
(4) $\boldsymbol{p}_{\boldsymbol{g}_s}^{\mathcal{B}}$ is the relative sub-goal position.
We represent $\bar{\boldsymbol{x}}_{\!\mathcal{M}}$ using a dual-map design, where the 3D map captures local volumetric occupancy for posture adjustment, while the 2.5D map provides a wider spatial context for navigation. 
This design improves training efficiency compared to a single 3D map $\boldsymbol{M}_{3\text{D}}^{\text{single}} \in \mathbb{R}^{31 \times 21 \times 11}$ at the same resolution, achieving comparable performance with approximately half wall-clock training time.
Then, a spatial-perception encoder $e_{\text{H}, \text{s}}^{\text{T}}$ maps $\bar{\boldsymbol{x}}_{\!\mathcal{M}}$ into a latent embedding $\boldsymbol{z}_{\text{s}} \in \mathbb{R}^{32}$, which is concatenated with other inputs and passed to the teacher's actor backbone
$b_{\text{H}}^{\text{T}}: \boldsymbol{z}_{\text{s}} \times \bar{\boldsymbol{o}}_{\text{p}} \times \boldsymbol{x}_{\text{m}} \times \boldsymbol{p}_{\boldsymbol{g}_s}^{\mathcal{B}} \rightarrow \boldsymbol{c}$.

The student policy $\pi_{\text{H}}^{\text{S}}$ is distilled from the fully-converged teacher policy.
It takes as input a 4-tuple observation $\bar{\boldsymbol{o}} = (\bar{\boldsymbol{o}}_{D}, \bar{\boldsymbol{o}}_{\text{p}}, \hat{\boldsymbol{x}}_{\text{m}}, \boldsymbol{p}_{\boldsymbol{g}}^{\mathcal{B}})$, encompassing
(1) $\bar{\boldsymbol{o}}_{D} \!\in \!\mathbb{R}^{180 \times 320}$, a depth image captured by an onboard front-facing camera; 
(2) $\bar{\boldsymbol{o}}_{\text{p}}$ is the high-level proprioceptive observation;
(3) $\hat{\boldsymbol{x}}_{\text{m}}$, the motion state estimated by the low-level motion estimator $e_{\text{L}, \text{m}}^{\text{S}}$; and 
(4) $\boldsymbol{p}_{\boldsymbol{g}}^{\mathcal{B}}$, the relative goal information.
A spatial-perception estimator $e_{\text{H}, \text{s}}^{\text{S}}$ encodes $\bar{\boldsymbol{o}}_{D}$ into the latent vector $\hat{\boldsymbol{z}}_{\text{s}}$ using internal recurrent memory, which is also fed to the student's backbone with other elements: $b_{\text{H}}^{\text{S}}:  \hat{\boldsymbol{z}}_{\text{s}} \times  \bar{\boldsymbol{o}}_{\text{p}} \times \hat{\boldsymbol{x}}_{\text{m}}  \times \boldsymbol{p}_{\boldsymbol{g}}^{\mathcal{B}} \rightarrow \hat{\boldsymbol{c}}$.

\setlength{\tabcolsep}{5pt} 
\begin{table}[t!]
\centering
\caption{Domain Randomization Parameters}
\vspace{-2mm}
\renewcommand{\arraystretch}{1.15}
\begin{tabular}{c|c|c|c}
\hline
\textbf{Parameter} & \textbf{Train Distribution} & \textbf{Test Distribution} & \textbf{Unit} \\
\hline
Payload  
& $\mathcal{U}(0.0,\,3.0)$  
& $\mathcal{U}(0.0,\,5.0)$  
& \si{\kilogram} \\

CoM Offset  
& $\mathcal{U}(-0.1,\,0.1)^3$  
& $\mathcal{U}(-0.15,\,0.15)^3$  
& \si{\meter} \\

Friction Coeff.  
& $\mathcal{U}(0.7,\,1.2)$  
& $\mathcal{U}(0.4,\,1.5)$  
& -- \\

Restitution Coeff.  
& $\mathcal{U}(0.0,\,0.1)$  
& $\mathcal{U}(0.0,\,0.3)$  
& -- \\

Motor Torque Scale  
& $\mathcal{U}(0.9,\,1.1)^{12}$  
& $\mathcal{U}(0.8,\,1.3)^{12}$  
& -- \\
\hline
\end{tabular}
\begin{tablenotes}
\setlength{\itemindent}{-0.3cm}
\item[] \textbullet\; $\mathcal{U}(a,b)^c$: uniform distribution over $[a,b]$ in each of the $c$ dimensions.
\end{tablenotes}
\vspace{-6mm}
\label{table:dom-rnd}
\end{table}

\subsection{Implementation Details} \label{sec:Implemetation Details}
We use the Unitree Go1 robot~\cite{Unitree_Go1} and the Isaac Gym simulator~\cite{makoviychuk2021isaac} for large-scale parallel data collection. 
Experiments are conducted on a desktop with an Intel Core i7-13700 CPU and an NVIDIA RTX 4090 GPU. 
We briefly depict the network composition in Fig.~\ref{fig:framework}.
At each level, the teacher policy is optimized using Proximal Policy Optimization (PPO)~\cite{schulman2017proximal} and distilled into the student policy via the DAgger algorithm. 

The high-level policy $\pi_H$ runs at 10~Hz.
The teacher policy is trained for 18 hours with 1,024 simulated robots across six $20 \times 20$~\si{\meter} unstructured environments described in Sec.~\ref{sec:task-formulation}, followed by 6 hours of student training with 128 robots. 
We adopt A* search over a Generalized Voronoi Diagram~\cite{yang2024path} to generate intermediate sub-goals detailed in Sec.~\ref{sec:way}. 
TABLE~\ref{table:high-level-rwd} enumerates the high-level reward function $\bar{\mathcal{R}} = \bar{\mathcal{R}}^{H}_{\text{task}} + \bar{\mathcal{R}}^{H}_{\text{reg}}$. 
To minimize the perceptual sim-to-real gap, the simulated depth images $\bar{\boldsymbol{o}}_{D}$ are augmented with random erasing, Gaussian blur, and additive noise, and the camera pose is randomized by applying translational and pitch offsets, $\Delta \boldsymbol{p} \sim \mathcal{U}(-0.05~\SI{}{\meter},\,0.05~\SI{}{\meter})^3$ and  $\Delta \theta_{y} \sim \mathcal{U}(-2.5^{\circ},\,2.5^{\circ})$, to the nominal pose $\boldsymbol{p}_{\mathrm{Cam}}^{\mathcal{B}} = [0.25,\,0.0,\,0.12]^{\top}$\SI{}{\meter} with zero tilting, followed by clipping at 4~\SI{}{\meter} and normalization to $[0,1]$.

The low-level policy $\pi_L$ operates at 50~Hz and is trained for 12 hours using 4,096 robots on a rough terrain comprising stairs, discrete holes, slopes, and flat surfaces, followed by 2 hours of student training with 300 robots. 
To narrow the sim-to-real gap, we apply domain randomization to intrinsic robot parameters $\boldsymbol{x}_{\text{d}}$, as detailed in TABLE~\ref{table:dom-rnd}, and inject random body-force impulses every \(10\,\text{s}\). 
The command $\boldsymbol{c}$ is sampled from the ranges: $v_{x}^{c} \in [-1.5, 1.5]$~\SI{}{\meter/\second}, $v_{y}^{c} \in [-1.0, 1.0]$~\SI{}{\meter/\second}, $\omega_{z}^{c} \in [-1.5, 1.5]$~\SI{}{\radian/\second}, $h^{c} \in [0.1, 0.4]$~\SI{}{\meter}, and $\theta_{x}^{c} \in [-1.0, 1.0]$~\SI{}{\radian}. 
TABLE~\ref{table:low-level-rwd} details the low-level reward $\mathcal{R} = \mathcal{R}^{L}_{\text{task}} + \mathcal{R}^{L}_{\text{reg}}$.

\begin{table}[t!]
\caption{High-Level Navigation Policy Rewards $\bar{\mathcal{R}}$}
\label{table:high-level-rwd}
\vspace{-2mm}
\renewcommand{\arraystretch}{1.3}
\begin{tabularx}{\linewidth}{>{\raggedleft\arraybackslash}p{0.35\linewidth}|>{\arraybackslash}p{0.65\linewidth}}
    \Xhline{1\arrayrulewidth}
    \textbf{Reward Term} & \textbf{Expression} \\
    \hline
    \multicolumn{2}{c}{\textit{\textbf{Task Rewards}}: \raisebox{0.1ex}{$\bar{\mathcal{R}}_{\text{task}}^H = \sum_{k=1}^{3} r^H_k$}} \\
    \Xhline{1\arrayrulewidth}
    \text{Goal Arrival} ($r^H_1$) & \raisebox{0.1ex}{$w^{H}_{1} \; \mathds{1}_{\text{goal}}$} \\
    \hline
    \text{State Count} ($r^H_2$) & \raisebox{0.1ex}{$w^{H}_{2} \; 1/\sqrt{\eta(\smash{\boldsymbol{p}_{B,xy}})}$} \\
    \hline
    \text{Desired Speed} ($r^H_3$) & \raisebox{0.1ex}{$w^{H}_{3} \; \exp( -| v_{\text{des}} - \|\boldsymbol{v}^{\mathcal{B}}_{B, xy}\|_2 | / 0.3)$} \\
\end{tabularx}
\begin{tabularx}{\linewidth}{>{\raggedleft\arraybackslash}p{0.35\linewidth}|>{\arraybackslash}p{0.65\linewidth}}
    \Xhline{1\arrayrulewidth}
    \multicolumn{2}{c}{\hspace{-1.23cm}\textit{\textbf{Regularization Rewards}}: \raisebox{0.1ex}{$\bar{\mathcal{R}}_{\text{reg}}^H = \sum_{k=4}^{10} r^H_k$}} \\
    \Xhline{1\arrayrulewidth}
    \text{Command Rate} ($r^H_4$) & \raisebox{0.1ex}{$w^{H}_{4} \; \|\boldsymbol{c}_t - \boldsymbol{c}_{t-1}\|_2^2$} \\
    \hline
    \text{Smooth Command} ($r^H_5$) & \raisebox{0.1ex}{$w^{H}_{5} \; \|\boldsymbol{c}_t - 2\boldsymbol{c}_{t-1} + \boldsymbol{c}_{t-2}\|_2^2$} \\
    \hline
    \text{Tracking Error} ($r^H_{6}$) & \raisebox{0.1ex}{$w^{H}_{6} \; \|\boldsymbol{c} - [ v^{\mathcal{B}}_{B, x}, v^{\mathcal{B}}_{B, y}, \omega^{\mathcal{B}}_{B, z}, h_{B}, \theta^{\mathcal{W}}_{B, x}] \|_2$} \\
    \hline
    \text{Body Velocity} ($r^H_{7}$) & \raisebox{0.1ex}{$w^{H}_{7} \; (|v^{\mathcal{B}}_{B,z}|^{2} + \|\boldsymbol{\omega}^{\mathcal{B}}_{B, xy}\|_2^2)$} \\
    \hline
    \text{Nominal Posture} ($r^H_{8}$) & \raisebox{0.1ex}{$w^{H}_{8} \; \|\boldsymbol{q} - \boldsymbol{q}_{0}\|_2^{2}$} \\
    \hline
    \text{Command limit} ($r^H_{9}$) & \raisebox{0.1ex}{$w^{H}_{9} \; \mathds{1}_{\text{command limit}}$} \\
    \hline
    \text{Collision} ($r^H_{10}$) & \raisebox{0.1ex}{$w^{H}_{10} \; \mathds{1}_{\text{col}}$} \\ 
\end{tabularx}
\begin{tabularx}{\linewidth}{ >{\arraybackslash}X|
                              >{\arraybackslash}X}
  \Xhline{1\arrayrulewidth}
    \multicolumn{2}{c}{\hspace{-.4cm}\textit{\textbf{Reward Weights}}} \\
  \Xhline{1\arrayrulewidth} 
    \multicolumn{2}{c}{
        \begin{tabularx}{\linewidth}{>{\arraybackslash}X
                                     >{\arraybackslash}X
                                     >{\arraybackslash}X
                                     >{\arraybackslash}X
                                     >{\arraybackslash}X
                                     >{\arraybackslash}X}

         \raisebox{0.1ex}{{\hspace{-2mm}\scriptsize$w^{H}_{1}$=$5.0$}} & 
         \raisebox{0.1ex}{{\hspace{-2mm}\scriptsize$w^{H}_{2}$=$0.5$}} & 
         \raisebox{0.1ex}{{\hspace{-2mm}\scriptsize$w^{H}_{3}$=$0.25$}} & 
         \raisebox{0.1ex}{{\hspace{-2mm}\scriptsize$w^{H}_{4}$=-$0.1$}} & 
         \raisebox{0.1ex}{{\hspace{-2mm}\scriptsize$w^{H}_{5}$=-$0.1$}} \\
        
         \raisebox{0.1ex}{{\hspace{-2mm}\scriptsize$w^{H}_{6}$=-$0.2$}} & 
         \raisebox{0.1ex}{{\hspace{-2mm}\scriptsize$w^{H}_{7}$=-$0.1$}} &
         \raisebox{0.1ex}{{\hspace{-2mm}\scriptsize$w^{H}_{8}$=-$4e$-$2$}}& 
         \raisebox{0.1ex}{{\hspace{-2mm}\scriptsize$w^{H}_{9}$=-$2.5$}} & 
         \raisebox{0.1ex}{{\hspace{-2mm}\scriptsize$w^{H}_{10}$=-$2.5$}} \\
        \end{tabularx}
    } \\
  \Xhline{1\arrayrulewidth}
\end{tabularx}
\begin{tablenotes}
\setlength{\itemindent}{-0.3cm}
\item[] \textbullet \; $\mathds{1}_{\text{goal}}$ returns 1 if $\|\boldsymbol{p}_{\boldsymbol{g}_{s}}^{\mathcal{W}}-\boldsymbol{p}_{B}^{\mathcal{W}}\| < \SI{0.1}{\meter}$, 0 otherwise.
\vspace{0.1em}
\item[] \textbullet\; $\mathds{1}_{\text{col}}$ returns 1 if the robot collides with the environment, and 0 otherwise.
\item[] \textbullet\; $\mathds{1}_{\text{command limit}}$ returns 1 if any command exceeds its limit, 0 otherwise.
\item[] \textbullet\; $v_{\text{des}}$ is the desired linear speed, sampled uniformly from [$0.3$, $1.2$] \SI{}{\meter/\second}.
\item[] \textbullet\; $\eta(\smash{\boldsymbol{p}_{B,xy}})$ is the discretized-state visitation count for the current position.
\end{tablenotes}
\vspace{-0.1mm}
\end{table}

\begin{table}[t!]
\vspace{-2.5mm}
\caption{Low-Level Locomotion Policy Rewards $\mathcal{R}$}
\label{table:low-level-rwd}
\vspace{-2mm}
\renewcommand{\arraystretch}{1.3}
\begin{tabularx}{\linewidth}{>{\raggedleft\arraybackslash}p{0.35\linewidth}|>{\arraybackslash}p{0.65\linewidth}}
    \Xhline{1\arrayrulewidth}
    \multicolumn{2}{c}{\textit{\textbf{Task Rewards}}: \raisebox{0.1ex}{$\mathcal{R}_{\text{task}}^L = \sum_{k=1}^{4} r^L_k$}} \\
    \Xhline{1\arrayrulewidth}
    \text{Velocity Tracking} ($r^L_1$) & \raisebox{0.1ex}{$w^{L}_{1} \; \exp( - \|[v_{x}^{c}, v_{y}^{c}] - \boldsymbol{v}^{\mathcal{B}}_{B,xy}\|_2 / 0.25 )$} \\
    \hline
    \text{Yaw-Rate Tracking} ($r^L_2$) & \raisebox{0.1ex}{$w^{L}_{2} \; \exp (-|\omega_{z}^{c} - \omega^{\mathcal{B}}_{B, z}| / 0.25 )$} \\
    \hline
    \text{Height Tracking} ($r^L_3$) & \raisebox{0.1ex}{$w^{L}_{3} \; \exp( -|h^{c} - h_{B}| / 0.0025 )$} \\
    \hline
    \text{Roll Tracking} ($r^L_4$) & \raisebox{0.1ex}{$w^{L}_{4} \; \exp( -|\theta_{x}^{c} - \theta^{\mathcal{W}}_{B, x}| / 0.05 )$} \\
\end{tabularx}
\begin{tabularx}{\linewidth}{>{\raggedleft\arraybackslash}p{0.35\linewidth}|>{\arraybackslash}p{0.65\linewidth}}
    \Xhline{1\arrayrulewidth}
    \multicolumn{2}{c}{\hspace{-1.235cm}\textit{\textbf{Regularization Rewards}}: \raisebox{0.1ex}{$\mathcal{R}_{\text{reg}}^L = \sum_{k=5}^{12} r^L_k$}} \\
    \Xhline{1\arrayrulewidth}
    \text{Action Rate} ($r^L_{5}$) & \raisebox{0.1ex}{$w^{L}_{5} \; \|\boldsymbol{a}_t - \boldsymbol{a}_{t-1}\|_2^2$} \\
    \hline
    \text{Smooth Action} ($r^L_{6}$) & \raisebox{0.1ex}{$w^{L}_{6} \; \|\boldsymbol{a}_t - 2\boldsymbol{a}_{t-1} + \boldsymbol{a}_{t-2}\|_2^2$} \\
    \hline
    \text{Body Orientation} ($r^L_7$) & \raisebox{0.1ex}{$w^{L}_{7} \; | \theta^{\mathcal{W}}_{B, y} |$} \\
    \hline
    \text{Body Velocity} ($r^L_8$) & \raisebox{0.1ex}{$w^{L}_{8}  \; |v^{\mathcal{B}}_{B,z}|^{2} + w^{L}_{9} \; \|\boldsymbol{\omega}^{\mathcal{B}}_{B, xy}\|_2^{2}$} \\
    \hline
    \text{Smooth Joint} ($r^L_{9}$) & \raisebox{0.1ex}{$w^{L}_{10} \; \|\dot{\boldsymbol{q}}\|_2^2 + w^{L}_{11} \; \|\ddot{\boldsymbol{q}}\|_2^{2}$} \\
    \hline
    \text{Torque Usage} ($r^L_{10}$) & \raisebox{0.1ex}{$w^{L}_{12} \; \|\boldsymbol{\tau}\|_2^2$} \\
    \hline
    \text{Joint Limit} ($r^L_{11}$) & \raisebox{0.1ex}{$w^{L}_{13} \; \mathds{1}_{\text{joint limit}}$} \\
    \hline
    \text{Collision} ($r^L_{12}$) & \raisebox{0.1ex}{$w^{L}_{14} \; \mathds{1}_{\text{col}}$} \\
\end{tabularx}
\begin{tabularx}{\linewidth}{ >{\arraybackslash}X|
                              >{\arraybackslash}X}
  \Xhline{1\arrayrulewidth}
    \multicolumn{2}{c}{\hspace{-.4cm}\textit{\textbf{Reward Weights}}} \\
  \Xhline{1\arrayrulewidth} 
    \multicolumn{2}{c}{
        \begin{tabularx}{\linewidth}{>{\arraybackslash}X
                                     >{\arraybackslash}X
                                     >{\arraybackslash}X
                                     >{\arraybackslash}X
                                     >{\arraybackslash}X
                                     >{\arraybackslash}X
                                     >{\arraybackslash}X}
         \raisebox{0.1ex}{{\hspace{-2.5mm}\scriptsize$w^{L}_{1}$=$0.4$}} & 
         \raisebox{0.1ex}{{\hspace{-2.5mm}\scriptsize$w^{L}_{2}$=$0.2$}} & 
         \raisebox{0.1ex}{{\hspace{-2.5mm}\scriptsize$w^{L}_{3}$=$0.2$}} & 
         \raisebox{0.1ex}{{\hspace{-2.5mm}\scriptsize$w^{L}_{4}$=$0.2$}} & 
         \raisebox{0.1ex}{{\hspace{-2.5mm}\scriptsize$w^{L}_{5}$=-$1e$-$2$}} & 
         \raisebox{0.1ex}{{\hspace{-2.5mm}\scriptsize$w^{L}_{6}$=-$0.1$}} & 
         \raisebox{0.1ex}{{\hspace{-2.5mm}\scriptsize$w^{L}_{7}$=-$5e$-$1$}} \\
         
         \raisebox{0.1ex}{{\hspace{-2.5mm}\scriptsize$w^{L}_{8}$=-$0.2$}} & 
         \raisebox{0.1ex}{{\hspace{-2.5mm}\scriptsize$w^{L}_{9}$=-$1e$-$3$}} & 
         \raisebox{0.1ex}{{\hspace{-2.5mm}\scriptsize$w^{L}_{10}$=-$1e$-$4$}} & 
         \raisebox{0.1ex}{{\hspace{-2.5mm}\scriptsize$w^{L}_{11}$=-$2e$-$7$}} & 
         \raisebox{0.1ex}{{\hspace{-2.5mm}\scriptsize$w^{L}_{12}$=-$1e$-$4$}} & 
         \raisebox{0.1ex}{{\hspace{-2.5mm}\scriptsize$w^{L}_{13}$=-$10$}} & 
         \raisebox{0.1ex}{{\hspace{-2.5mm}\scriptsize$w^{L}_{14}$=-$10$}} 
        \end{tabularx}
    } \\
  \Xhline{1\arrayrulewidth}
\end{tabularx}
\begin{tablenotes}
\setlength{\itemindent}{-0.3cm}
\item[] \textbullet\; $\mathds{1}_{\text{col}}$ returns 1 if the robot collides with the environment, and 0 otherwise.
\item[] \textbullet\; $\mathds{1}_{\text{joint limit}}$ returns 1 if any joint exceeds its limit, 0 otherwise.
\end{tablenotes}
\vspace{-7mm}
\end{table}
\section{Experimental Results} \label{Sec:4}

We conducted extensive experiments in both simulation and real-world environments to validate the effectiveness of the proposed Hierarchical Posture-Adaptive Navigation (HiPAN) framework. 
We designed the experiments to answer three key questions: 
(1) Does the posture adaptation improve navigation efficiency in unstructured 3D environments? 
(2) How effective is path-guided curriculum learning for long-horizon navigation? 
(3) Can the framework trained in simulation transfer robustly to physical quadruped robots using only onboard perception? 

\begin{table*}[!ht]
\centering
\renewcommand{\arraystretch}{0.855}
\caption{Navigation Performance in Simulation Environments. For Details of Experimental Setups, Please Refer to Sec.~\ref{sec:eval_env} \& Sec.~\ref{sec:eval_baselines}}
\vspace{-2mm}
\label{tab:navigation_performance}
\begin{tabular}{clc@{\hspace{0.10em}}ccc@{\hspace{0.05em}}ccc@{\hspace{0.05em}}ccc@{\hspace{0.05em}}cc}
\toprule 
\addlinespace[0.6mm]
& & & \multicolumn{11}{c}{\fontsize{8pt}{8pt}\selectfont\textbf{Environment Complexity $( \text{Low} \rightarrow \text{High} )$}}\\ [-0.05ex]
\cmidrule(r){4-14}
\multirow{2}{*}{\textbf{Groups}}& \multirow{2}{*}{\textbf{Methods}}& & \multicolumn{2}{c}{\textit{Corridor}}& &\multicolumn{2}{c}{\textit{Room}} & &\multicolumn{2}{c}{\textit{Complex-1}}& &\multicolumn{2}{c}{\textit{Complex-2}}\\
\cmidrule(r){4-5} \cmidrule(r){7-8} \cmidrule(r){10-11} \cmidrule(r){13-14}
 & & &  SR($\%$ , $\uparrow$) & SPL(- , $\uparrow$) & &  SR($\%$ , $\uparrow$) & SPL(- , $\uparrow$) & &  SR($\%$ , $\uparrow$) & SPL(- , $\uparrow$) & &  SR($\%$ , $\uparrow$) & SPL(- , $\uparrow$)\\
\midrule
- & HiPAN (\textit{Ours}) (\textit{Sec.~\ref{sec:method})}  & & 98.5$\pm$1.1 & 93.2$\pm$1.3 & & \textbf{98.4$\pm$0.9} & \textbf{88.8$\pm$1.2} & & \textbf{94.4$\pm$1.0} & \textbf{89.3$\pm$1.2} & & \textbf{94.7$\pm$1.3}  & \textbf{83.6$\pm$1.2}\\
\midrule
\multirow{3}{*}{\textit{G1}} & HiPAN w/o Pos. Adjustment & & 98.4$\pm$0.8 & 93.1$\pm$1.0 & & 94.5$\pm$1.4 & 74.1$\pm$1.3 & & 74.2$\pm$1.8 & 69.1$\pm$2.0 & & 73.1$\pm$1.6 & 66.2$\pm$2.2\\
& \textit{Bug}~\cite{mcguire2019comparative, zohaib2014improved, abafogi2018new}          & & 91.1$\pm$1.9 & 58.9$\pm$0.8 & & 88.1$\pm$1.2 & 67.4$\pm$1.2 & & 23.8$\pm$1.4 & 20.1$\pm$1.2 & & 21.4$\pm$1.5 & 20.4$\pm$1.4\\
& \textit{Wall-Following}~\cite{alamri2023autonomous}    & & 51.9$\pm$2.6 & 31.7$\pm$2.8 & & 84.5$\pm$1.3 & 59.1$\pm$1.4 & & 22.4$\pm$1.5 & 19.8$\pm$1.6 & & 20.2$\pm$1.4 & 19.7$\pm$1.1\\
\midrule
\multirow{3}{*}{\textit{G2}} & HiPAN w/o IR & & \textbf{99.0$\pm$0.3} & \textbf{93.2$\pm$0.8} & & 96.2$\pm$1.5 & 86.2$\pm$2.0 & & 90.8$\pm$1.4 & 81.3$\pm$0.8 & & 90.7$\pm$1.8 & 82.1$\pm$1.4\\
& HiPAN w/o PGCL & & 97.8$\pm$1.4 & 88.7$\pm$1.3 & & 76.9$\pm$1.8 & 69.0$\pm$1.5 & & 78.1$\pm$1.8 & 68.0$\pm$1.4 & & 74.5$\pm$1.8 & 66.5$\pm$1.7\\
& HiPAN w/o IR, PGCL & & 96.6$\pm$1.4 & 87.3$\pm$1.0 & & 53.0$\pm$0.8 & 43.5$\pm$1.9 & & 50.5$\pm$1.6 & 42.8$\pm$1.3 & & 50.1$\pm$1.3 & 43.6$\pm$1.2\\
\midrule
\textit{G3} & \textit{Flat-RL}~\cite{rudin2022advanced}  & & 82.0$\pm$1.4 & 71.5$\pm$0.9 & & 43.8$\pm$5.6 & 39.7$\pm$4.4 & & 45.3$\pm$1.0 & 42.1$\pm$0.8 & & 43.7$\pm$0.8 & 40.1$\pm$1.1 \\
\midrule
\multirow{4}{*}{\textit{G4}} & HiPAN w/ \textit{RPL}~\cite{zhuang2023robot} & & 98.3$\pm$1.3 &93.2$\pm$1.3 & & 98.2$\pm$0.8 & 86.8$\pm$1.2 & & 90.2$\pm$1.4 & 85.6$\pm$1.4 & & 90.6$\pm$1.5 & 80.5$\pm$1.6\\
& $\hookrightarrow$ w/o PGCL & & 97.7$\pm$1.3 &88.7$\pm$1.4 & & 76.2$\pm$1.7 & 68.8$\pm$1.4 & & 75.7$\pm$1.5 & 66.4$\pm$1.6 & & 71.4$\pm$1.7 & 65.3$\pm$1.5\\
& HiPAN w/ \textit{MoE-Loco}~\cite{huang2025moe} & & 98.4$\pm$1.3 &93.1$\pm$1.2 & & 97.0$\pm$1.4 & 88.5$\pm$1.4 & & 85.3$\pm$1.6 & 80.1$\pm$1.7 & & 85.5$\pm$1.8 & 75.6$\pm$1.6\\
& $\hookrightarrow$ w/o PGCL & & 97.2$\pm$1.4 &88.0$\pm$1.4 & & 76.0$\pm$1.9 & 68.7$\pm$1.6 & & 70.7$\pm$1.8 & 62.3$\pm$1.7 & & 67.5$\pm$1.8 & 61.5$\pm$1.7 \\
\bottomrule
\end{tabular}
\vspace{-4.0mm}
\end{table*}

\subsection{Benchmark Environments and Performance Metrics} \label{sec:eval_env}
\begin{figure}[t!]
\centering
\vspace{-1mm}
\includegraphics[width=1.0\linewidth]{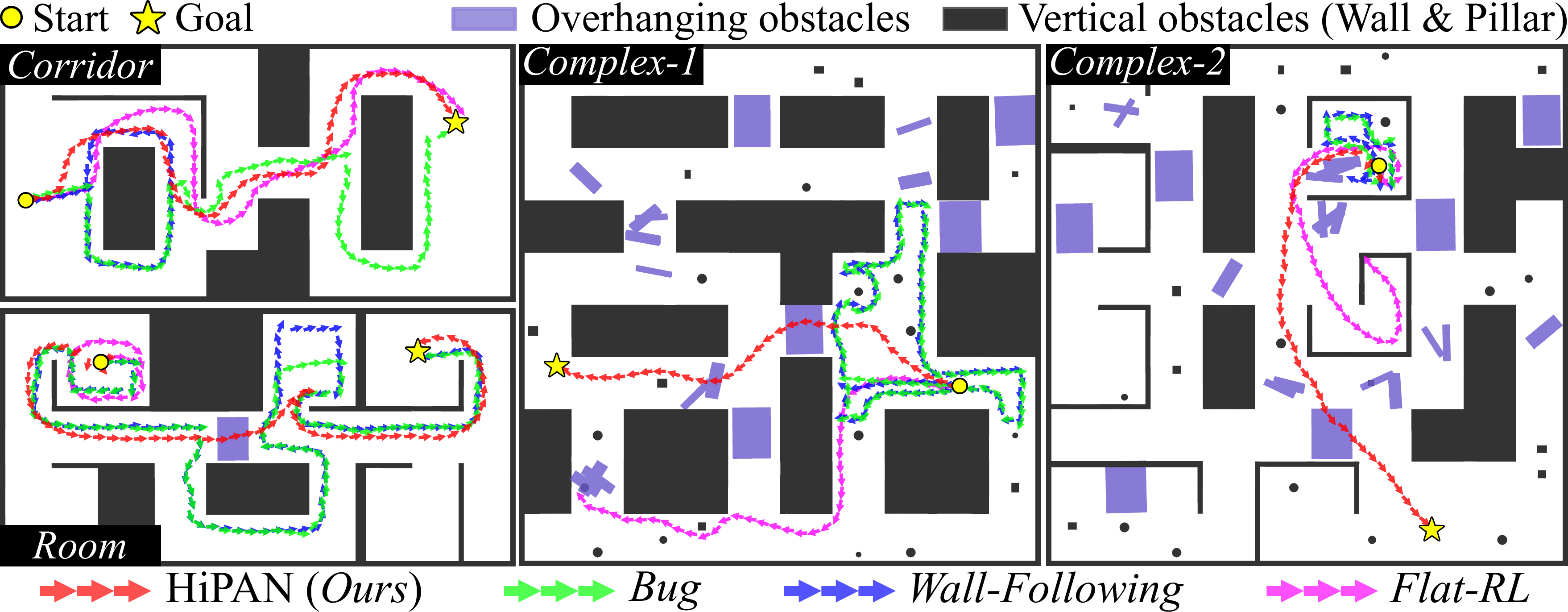}
    \vspace{-7mm}
    \caption{
    \textbf{Four benchmark environments and qualitative navigation results.}
    For each environment, trajectories are visualized as discrete arrows sampled every \SI{0.5}{\meter}, indicating the robot’s position and heading orientation under a fixed start–goal setting.
    The proposed HiPAN (red) reliably reaches the goal, whereas \textit{Bug} (green), \textit{Wall-Following} (blue), and \textit{Flat-RL} (magenta) often take inefficient detours or become trapped, failing to complete the task.
    } 
    \label{fig:5}
    \vspace{-6mm}
\end{figure}

We evaluated our framework in procedurally generated unstructured 3D environments, encompassing dead-end corridors, semi-enclosed rooms, and complex layouts with overhanging obstacles. 
Fig.~\ref{fig:5} illustrates four benchmarking environments: \textit{Corridor}, \textit{Room}, \textit{Complex-1}, and \textit{Complex-2}.
\textit{Corridor} examines the agent’s basic ability to navigate towards the goal when the direct path is obstructed by walls.
\textit{Room} evaluates the ability to employ posture adjustment for efficient traversal under overhanging obstacles---placed at the center of the map---and to escape semi-enclosed spaces using a non-greedy navigation strategy.
\textit{Complex-1} and \textit{Complex-2} assess overall navigation performance in unstructured 3D environments, requiring posture adaptation and long-horizon traversal through densely cluttered obstacles and intricate layouts.
For each environment, we generated $300$ navigation tasks by randomly sampling $100$ collision-free start–goal pairs for each distance range: $[5, 10]$, $[10, 20]$, and $[20, 30]$~\si{\meter}.
Mean and standard deviation are measured over 10 different seeds.

We employ two standard metrics for quantitative evaluation:
Success Rate (SR) assesses the proportion of navigation tasks in which the agent reaches the goal within a \SI{0.2}{\meter} threshold, while Success weighted by Path Length (SPL)~\cite{anderson2018evaluation} quantifies path efficiency as the success rate weighted by the ratio of the shortest feasible to the actual traversed path length~\cite{yang2024path}.

\subsection{Baselines and Ablation Methods} \label{sec:eval_baselines}
We organize a variety of baselines and ablated variants of HiPAN (\textit{Ours}) into four groups, as enumerated in TABLE~\ref{tab:navigation_performance}.
Group \textit{G1} consists of classical local navigation strategies, \textit{Bug}-Algorithm~\cite{mcguire2019comparative, zohaib2014improved, abafogi2018new} and \textit{Wall-Following}~\cite{alamri2023autonomous}, along with a HiPAN variant, HiPAN w/o Pos. Adjustment, that omits the postural commands $(h^{c}, \theta_{x}^{c})$ from the navigation command $\boldsymbol{c}$.
This group compares classical rule-based and learning-based navigation strategies excluding posture adjustment and also highlights the role of posture adaptation through comparison with HiPAN (\textit{Ours}).
\textit{G2} comprises HiPAN variants, designed to evaluate the contributions of the Intrinsic Reward (IR) and Path-Guided Curriculum Learning (PGCL) introduced in Sec.~\ref{sec:way}. 
\textit{G3} encompasses a \textit{Flat-RL} baseline~\cite{rudin2022advanced}, which employs a single end-to-end policy mapping observations directly to joint actions.
This baseline serves to validate the effectiveness of our overall system architecture against the alternative end-to-end approach without hierarchy.
Finally, \textit{G4} presents implicit posture-adaptive locomotion policies as low-level controllers within our framework, where the high-level policy outputs only velocity commands and posture is autonomously adapted at the low level using either exteroceptive perception (HiPAN w/ \textit{RPL}~\cite{zhuang2023robot}) or solely proprioception (HiPAN w/ \textit{MoE-Loco}~\cite{huang2025moe}), to examine alternative design choices for posture adaptation.

\vspace{-1.0mm}
\subsection{Evaluation Results} \label{sec:eval_results}
TABLE~\ref{tab:navigation_performance} shows the quantitative results. 
HiPAN (\textit{Ours}) achieves the highest success rate (SR) and path efficiency (SPL), outperforming the classical baselines in \textit{G1} and the end-to-end RL policy in \textit{G3}.
Fig.~\ref{fig:5} provides qualitative results on representative scenarios, showing that \textit{Ours} reliably reaches the goal, whereas baselines often take detours or become trapped.

Among \textit{G1} baselines, HiPAN w/o Pos. Adjustment surpasses classical \textit{Bug} and \textit{Wall-Following} methods by achieving higher success rates and more efficient paths. 
However, as shown in the \textit{Room}, the absence of posture adaptation leads to lower SPL compared to \textit{Ours} due to detours around overhanging obstacles.
Moreover, as illustrated in Fig.~\ref{fig:5}, classical methods deteriorate in \textit{Complex-1} and \textit{-2}, where dense obstacles and irregular layouts often entrap the agent.

Furthermore, we examine the contributions of the Intrinsic Reward (IR) and Path-Guided Curriculum Learning (PGCL) to navigation performance.
IR promotes state exploration by providing auxiliary incentives to novel states. 
While PGCL reshapes the task into a sequence of subproblems aligned with agent proficiency, providing structured guidance for effective exploration.
As shown in \textit{G2} in TABLE~\ref{tab:navigation_performance}, removing PGCL leads to greater performance degradation than removing IR, indicating that structured guidance is particularly beneficial in unstructured environments. 
Moreover, these two methods are complementary: IR provides a general incentive to visit any novel state, while PGCL prioritizes task-relevant states. 
This synergy allows \textit{Ours} to achieve superior performance in all environments except the simple \textit{Corridor} environment.

The \textit{Flat-RL} in \textit{G3} suffers from a high learning burden, as it must learn locomotion, posture adjustment, and navigation strategies simultaneously.
Early in training, traversing confined spaces with immature locomotion leads to frequent collisions and penalties, causing \textit{Flat-RL} to develop a bias for detouring.
This results in less efficient paths and causes the agent to get stuck if a detour is unavailable, as shown in \textit{Complex-1} of Fig.~\ref{fig:5}.
Furthermore, its myopic navigation strategy prioritizing goal proximity prevents the agent from escaping local minima, causing it to remain trapped as depicted in \textit{Room} of Fig.~\ref{fig:5}.
Consequently, its performance is markedly lower than \textit{Ours}.

\begin{figure*}[t]
\centering
\includegraphics[width=\textwidth]{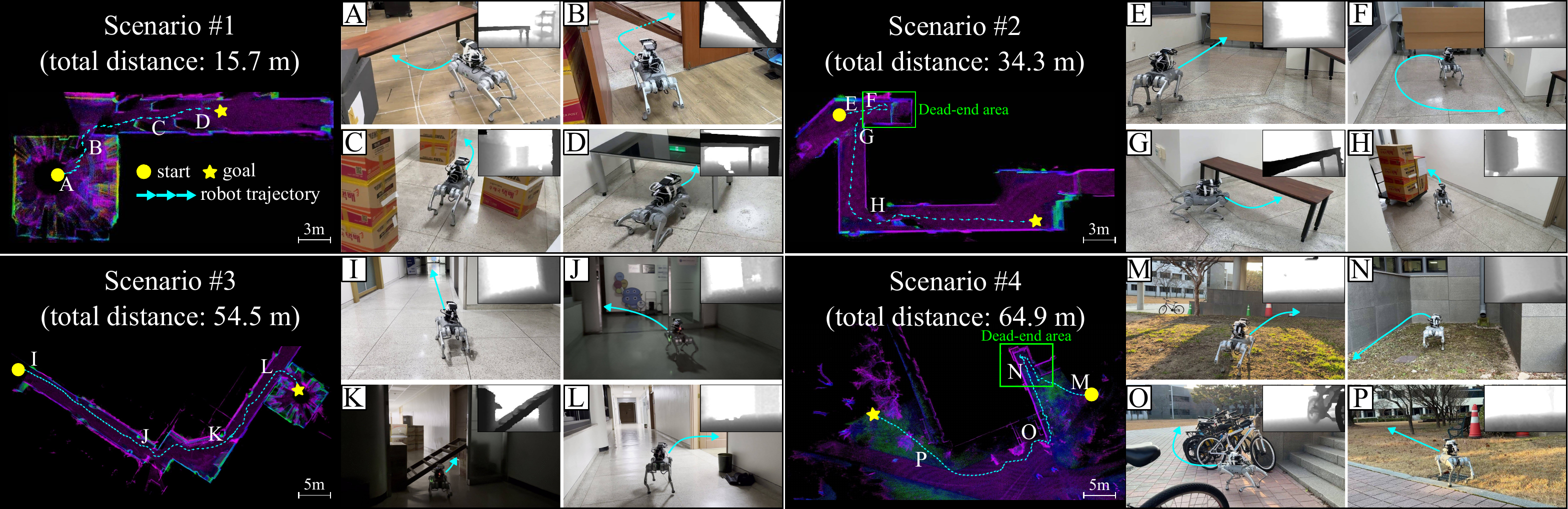} 
\vspace{-7mm}
\caption{
\textbf{Real-world navigation experiments in unstructured environments.}
For each scenario, the left panel shows the robot trajectory (cyan arrows) from the start (circle) to the goal (star), overlaid on LiDAR-based reconstructed maps provided solely for visualization. The right panels present representative robot behaviors with depth-image insets. Scenario~1 demonstrates posture-adaptive navigation in a cluttered indoor environment. Scenario~2 highlights recovery from local minima by backtracking from a dead-end corridor and continuing toward the goal. Scenario~3 validates robustness to abrupt illumination changes. Scenario~4 illustrates outdoor navigation under direct sunlight, including dead-end recovery and avoidance of diverse obstacles not encountered during training.
}
\label{fig:real experiments}
\vspace{-5mm}
\end{figure*}

\begin{figure}[t!]
\centering
\includegraphics[width=0.9\linewidth]{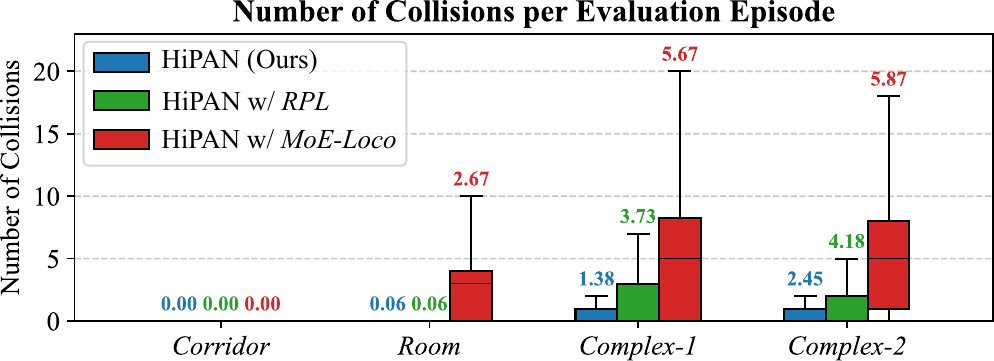}
\vspace{-3mm}
\caption{
\textbf{Collision statistics across benchmark environments.} Box plots compare HiPAN (\textit{Ours}) with the \textit{G4} baselines listed in Sec.~\ref{sec:eval_baselines}, showing per-episode collision distributions; numerical labels denote the mean values.
}
\label{fig:7}
\vspace{-5.5mm}
\end{figure}

Lastly, HiPAN w/ \textit{RPL} and w/ \textit{MoE-Loco} in \textit{G4} demonstrate slightly degraded performance compared to \textit{Ours}. Their reliance on a prefixed low-level policy prevents joint exploration of velocity and posture at the navigation level, constraining the solution space for traversing irregular obstacles and narrow passages. 
HiPAN w/ \textit{RPL} further experiences distributional shift when high-level velocity commands drive the agent into states insufficiently covered by the low-level controller. 
Meanwhile, HiPAN w/ \textit{MoE-Loco}, which relies solely on proprioception for posture adaptation, requires physical contact to trigger adjustment. These factors lead to increased collisions, as shown in Fig.~\ref{fig:7}, indicating reduced stability.
Importantly, orthogonal to the choice of low-level baselines, our PGCL remains essential for effective long-horizon navigation by promoting a well-balanced exploration–exploitation behavior.

\vspace{-2.0mm}
\subsection{Real-World Experiments} \label{sec:real}
We validated HiPAN on a Unitree Go1 robot~\cite{Unitree_Go1} equipped with a RealSense D435i camera, using an Intel NUC for onboard computation and ROS for inter-module communication. To mitigate the domain gaps described in Sec.~\ref{sec:Implemetation Details}, the captured depth images $\bar{\boldsymbol{o}}_{D}$ were clipped, normalized, and processed with a hole-filling filter to suppress sensor noise. 

Fig.~\ref{fig:real experiments} illustrates four real-world navigation experiments.
For indoor evaluations (Scenarios~1--3), we constructed unstructured 3D environments by randomly placing tables, ladders, and box piles in spaces with enclosed rooms, doors, and dead-end corridors.
Scenario~4 extends the evaluation to an outdoor environment featuring direct sunlight, uneven and sloped terrain, and obstacles unseen during training (e.g., bicycles and traffic cones).
These scenarios require recovery from local minima, posture adaptation, and robustness to lighting variations.
Despite these challenges, HiPAN successfully guided the robot to the goal position.
The supplementary video provides an intuitive visual understanding of these navigation examples in unstructured 3D environments.

\section{Conclusion and Future Work} \label{sec:6}
We introduce the Hierarchical Posture-Adaptive Navigation (HiPAN) framework, which enables quadruped robots to robustly navigate unstructured 3D environments relying solely on onboard depth perception without explicit map reconstruction. Simulation experiments show that HiPAN consistently outperforms conventional and end-to-end baselines in both success rate and path efficiency, while ablation studies confirm the contributions of the Path-Guided Curriculum Learning (PGCL) and the hierarchical system design with the explicit navigation command for safe and long-range navigation.
Real-world experiments further validate the practical applicability of HiPAN across diverse unstructured environments.

However, during real-world deployments, we observed that transparent or specular surfaces (e.g., glass walls and doors) are challenging to perceive accurately using infrared/stereo-based depth sensors, which may cause the robot to misinterpret them as traversable space. As future work, we plan to incorporate semantic awareness to recognize transparent obstacles and signage, including exit indicators, thereby further improving navigation effectiveness and robustness.

\bibliographystyle{IEEEtran}
\bibliography{IEEEabrv, bibliography.bib}

\end{document}